\RequirePackage{snapshot}
\documentclass[conference]{IEEEtran}

\usepackage{times}
\usepackage{amsfonts,amsmath,amssymb}
\usepackage{flushend}
\usepackage{bigints}
\usepackage{hhline}

\usepackage{graphicx}
\usepackage{booktabs, multicol, multirow}
\usepackage{indentfirst}
\usepackage{subfigure}

\usepackage{caption}

\usepackage{marginnote}
\usepackage[usenames,dvipsnames]{color}
\definecolor{fullred}{rgb}{0.85,.0,.1}

\usepackage[textsize=tiny]{todonotes}

\usepackage{array}
\newcolumntype{+}{>{\global\let\currentrowstyle\relax}}
\newcolumntype{^}{>{\currentrowstyle}}

\newcolumntype{C}[1]{>{\centering\arraybackslash}p{#1}}

\usepackage[numbers]{natbib}
\usepackage{multicol}
\usepackage[bookmarks=true]{hyperref}

\usepackage{algorithm,algpseudocode}
\algnewcommand{\Inputs}[1]{%
  \vspace{1mm}
  \State \textbf{Inputs:}
  \Statex \hspace*{\algorithmicindent}\parbox[t]{.8\linewidth}{\raggedright #1}
}
\algnewcommand{\Initialize}[1]{%
  \vspace{1mm}
  \State \textbf{Initialize:}
  \Statex \hspace*{\algorithmicindent}\parbox[t]{.8\linewidth}{\raggedright #1}
}
\algnewcommand{\Outputs}[1]{%
  \vspace{1mm}
  \State \textbf{Outputs:}
  \Statex \hspace*{\algorithmicindent}\parbox[t]{.8\linewidth}{\raggedright #1}
}
\algnewcommand{\Notation}[1]{%
  \vspace{1mm}
  \State \textbf{Notation:}
  \Statex \hspace*{\algorithmicindent}\parbox[t]{.8\linewidth}{\raggedright #1}
}
\algnewcommand{\Proc}[1]{%
  \vspace{1mm}
  \State \textbf{Procedure:}
  \Statex \hspace*{\algorithmicindent}\parbox[t]{.8\linewidth}{\raggedright #1}
}

\ifx\pdfoutput\undefined
 
\else
\ifx\pdfoutput\relax
 
\else
\ifnum\pdfoutput>0
\else
 
\fi
\fi
\fi

\begin{document}

\title{Learning Articulated Motions\\From Visual Demonstration}

\author{\authorblockN{Sudeep Pillai,
Matthew R.~Walter and
Seth Teller}
\authorblockA{Computer Science and Artificial Intelligence Laboratory\\
Massachusetts Institute of Technology\\
Cambridge, MA 02139 USA\\
Email: \{spillai, mwalter, teller\}@csail.mit.edu}}



\maketitle

\begin{abstract}
Many functional elements of human homes and workplaces consist of
rigid components which are connected through one or more sliding
or rotating linkages.  Examples include doors and drawers of
cabinets and appliances; laptops; and swivel office chairs.  A robotic
mobile manipulator would benefit from the ability to acquire kinematic
models of such objects from observation. This paper describes a method
by which a robot can acquire an object model by capturing depth
imagery of the object as a human moves it through its range of motion.
We envision that in future, a machine newly introduced to an
environment could be shown by its human user the articulated
objects particular to that environment, inferring from these
``visual demonstrations'' enough information to actuate each
object independently of the user.

Our method employs sparse (markerless) feature tracking, motion
segmentation, component pose estimation, and articulation learning; it
does not require prior object models.  Using the method, a robot can
observe an object being exercised, infer a kinematic model
incorporating rigid, prismatic and revolute joints, then use the model
to predict the object's motion from a novel vantage point.  We
evaluate the method's performance, and compare it to that of a
previously published technique, for a variety of household objects.

\end{abstract}

\IEEEpeerreviewmaketitle

\section{Introduction}

A long-standing challenge in robotics is to endow robots with the
ability to interact effectively with the diversity of objects common
in human-made environments. Existing approaches to manipulation often
assume that objects are simple and drawn from a small set. The models
are then either pre-defined or learned from training, for example
requiring fiducial markers on object parts, or prior assumptions
about object structure.  Such requirements may not scale well as
the number and variety of objects increases.  This paper describes
a method with which robots can learn kinematic models for articulated 
objects in situ, simply by observing a user manipulate the object.
Our method learns open kinematic chains that involve rigid
linkages, and prismatic and revolute motions, between parts.


%
\begin{figure}[!t]
  \centering
  \begin{tabular}{c}
    \includegraphics[width=\columnwidth]{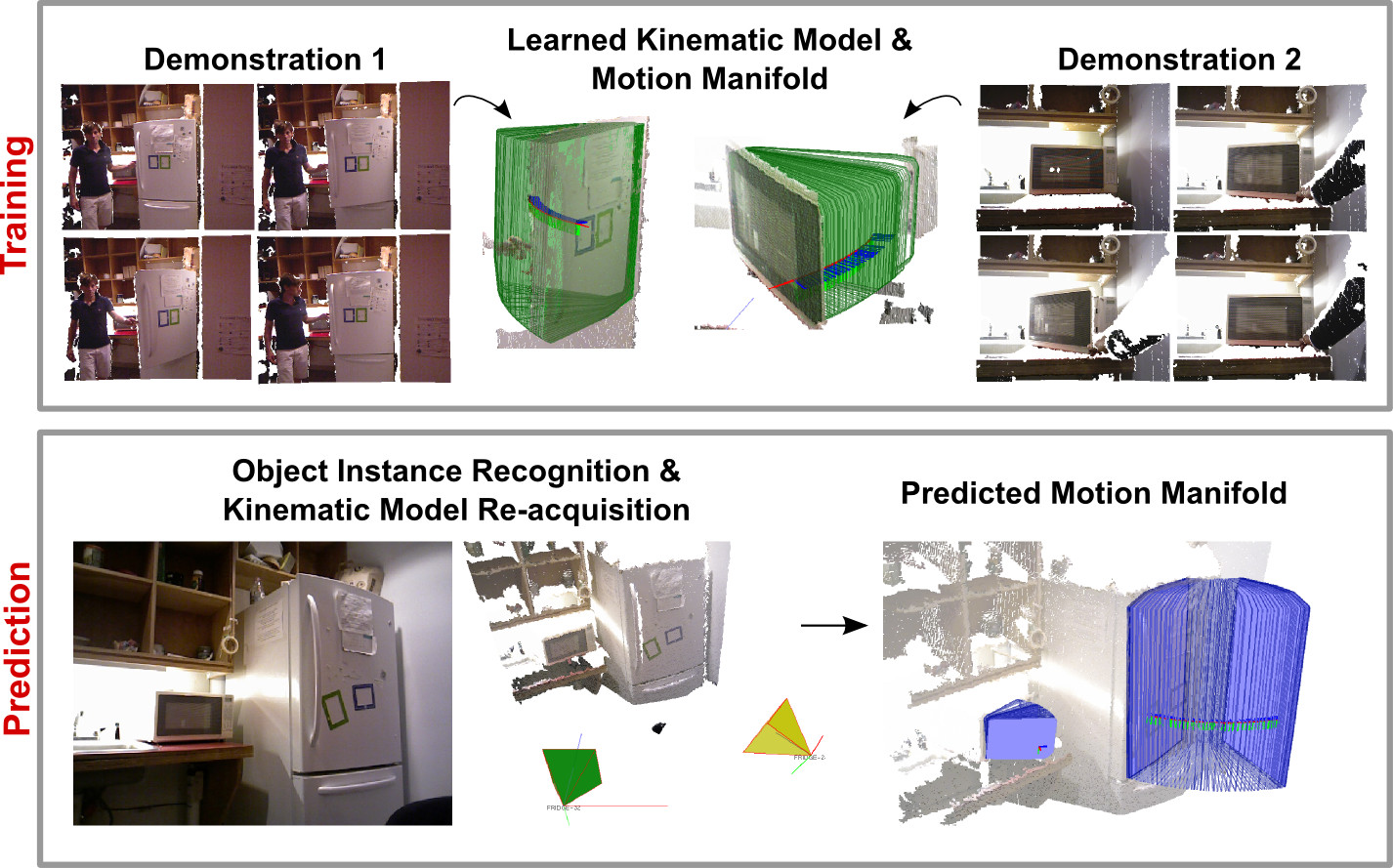}\\
  \end{tabular}
  \caption{The proposed framework reliably learns the
    underlying kinematic model of multiple articulated objects from
    user-provided visual demonstrations, and subsequently predicts their motions
    at future encounters.}
  \label{fig:intro-fig}
\end{figure}


There are three primary contributions of our approach that make it
effective for articulation learning. First, we propose a feature
tracking algorithm designed to perceive articulated motions in
unstructured environments, avoiding the need to embed fiducial markers
in the scene.  Second, we describe a motion segmentation
algorithm that uses kernel-based clustering to group feature trajectories arising
from each object part. A subsequent optimization step recovers the
6-DOF pose of each object part. Third, the method enables use of the
learned articulation model to predict the object's motion when it is
observed from a novel vantage point. Figure~\ref{fig:intro-fig}
illustrates a scenario where our method learns kinematic models for a
refrigerator and microwave from separate user-provided demonstrations,
then predicts the motion of each object in a subsequent encounter.  We
present experimental results that demonstrate the use of our method to
learn kinematic models for a variety of everyday objects, and compare
our method's performance to that of the current state of the art.


%
\begin{figure*}[!t]
    \centering
    \includegraphics[width=2\columnwidth]{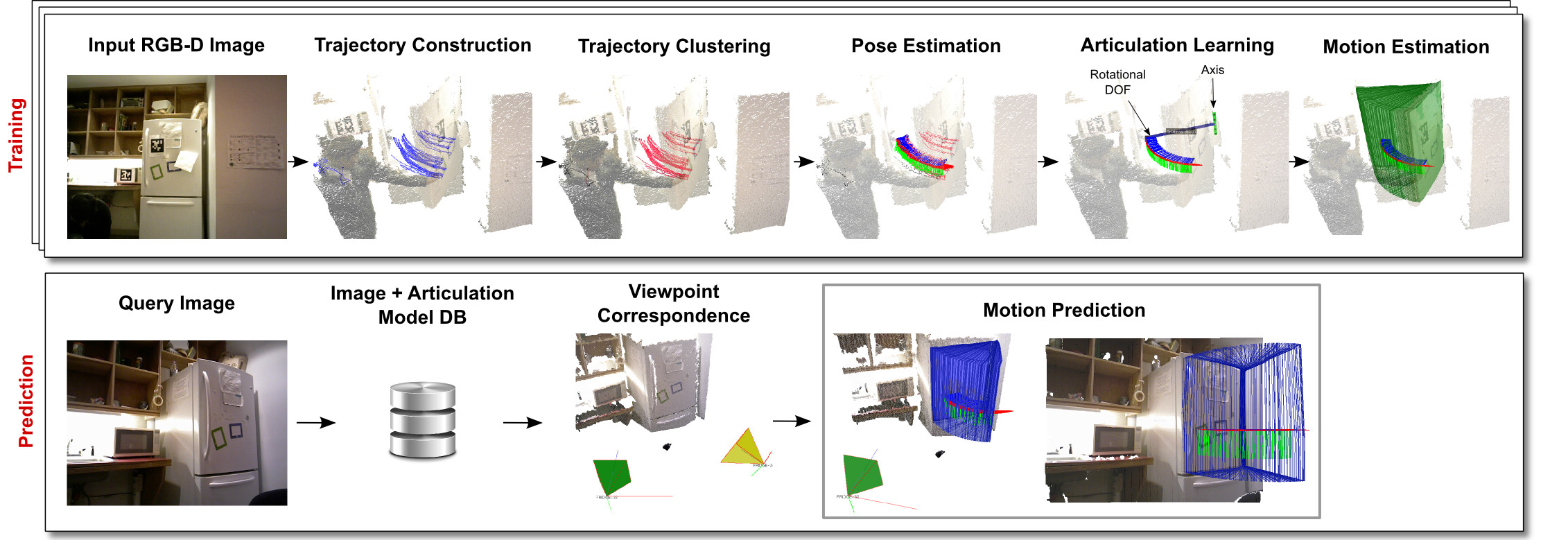}
    \caption{Articulation learning architecture.}
    \label{fig:vis-demonstration-pipeline}
\end{figure*}
\section{Related Work}
\label{sec:related-work}
Providing robots with the ability to learn models of articulated
objects requires a range of perceptual skills such as object tracking,
motion segmentation, pose estimation, and model learning. It is
desirable for robots to learn these models from demonstrations
provided by ordinary users.  This necessitates the ability to deal
with unstructured environments and estimate object motion without
requiring tracking markers. Traditional tracking algorithms such as
KLT~\cite{bouguet2001pyramidal}, or those based on
SIFT~\cite{lowe2004distinctive} depend on sufficient object texture
and may be susceptible to drift when employed over an object's full
range of motion. Alternatives such as large-displacement optical
flow~\cite{brox2011large} or particle video
methods~\cite{sand2008particle} tend to be more accurate but require
substantially more computation.

Articulated motion understanding generally requires a combination of motion
tracking and segmentation. Existing motion segmentation algorithms use feature
based trackers to construct spatio-temporal trajectories from sensor data, and
cluster these trajectories based on rigid-body motion constraints.  Recent work
by ~\citet{brox2010object} in segmenting feature trajectories has shown promise
in analyzing and labeling motion profiles of objects in video sequences in an
unsupervised manner. Recent work by~\citet{elhamifar2009sparse} has proven
effective at labeling object points based purely on motion visible in a sequence
of standard camera images. Our framework employs similar techniques, and
introduce a segmentation approach for features extracted from RGB-D data.

Researchers have studied the problem of learning models from visual
demonstration. \citet{yan2006general} and \citet{huang12} employ structure from
motion techniques to segment the articulated parts of an object, then estimate
the prismatic and rotational degrees of freedom between these parts. These
methods are sensitive to outliers in the feature matching step, resulting in
significant errors in pose and model estimates. Closely related to our work,
\citet{katz2010interactive} consider the problem of extracting segmentation and
kinematic models from interactive manipulation of an articulated object. They
take a deterministic approach, first assuming that each object linkage is
prismatic and proceed to fit a rotational degree-of-freedom only if the residual
is above a specified threshold. Katz et al.\ learn from observations made in
clean, clutter-free environments and primarily consider objects in close
proximity to the RGB-D sensor. Recently, \citet{katz2013interactive} propose an
improved learning method that has equally good performance with reduced
algorithmic complexity. However, the method does not explicitly reason over the
complexity of the inferred kinematic models, and tends to over-fit to observed
motion. In contrast, our algorithm targets in situ learning in unstructured
environments with probabilistic techniques that provide robustness to noise. Our
method adopts the work of \citet{sturm2011probabilistic}, which used a
probabilistic approach to reason over the likelihood of the observations while
simultaneously penalizing complexity in the kinematic model. Their work differs
from ours in two main respects: they required that fiducial markers be placed on
each object part in order to provide nearly noise-free observations; and they
assume that the number of unique object parts is known a priori.

\section{Articulation Learning From Visual Demonstration}
\label{sec:procedure}
This section introduces the algorithmic components of our method.
Figure~\ref{fig:vis-demonstration-pipeline} illustrates the
steps involved.

\begin{figure}[h]
  \centering
    \includegraphics[width=0.95\columnwidth]{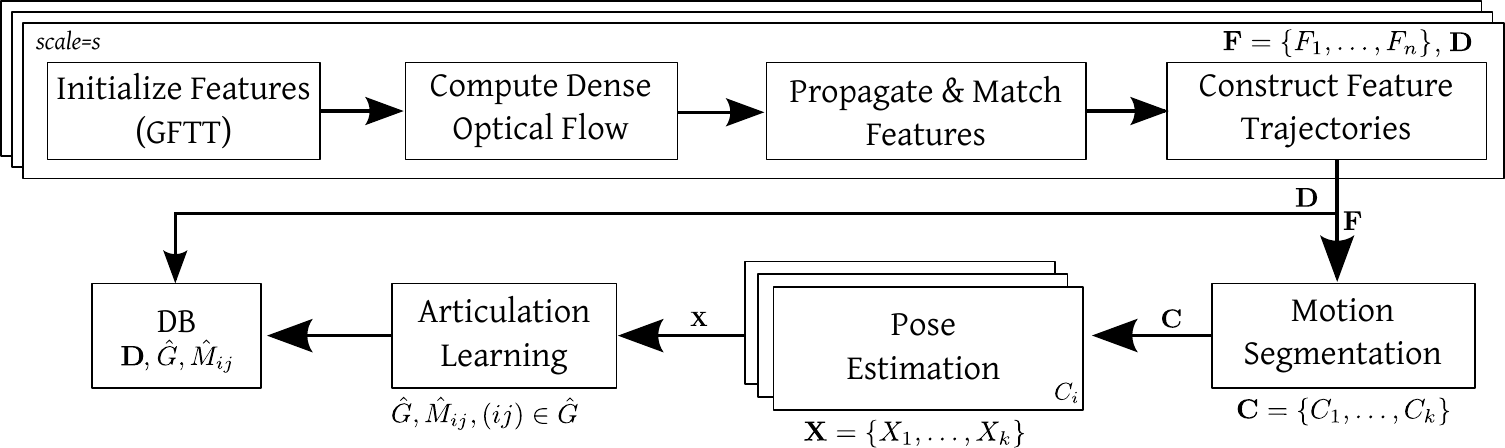}
  \caption{The training phase.}
  \label{fig:pipeline-text-training}
\end{figure}

Our approach consists of a training phase and a prediction phase. The
training phase proceeds as follows: (i) Given RGB-D data, a feature
tracker constructs long-range feature trajectories in 3-D. (ii) Using
a relative motion similarity metric, clusters of rigidly moving
feature trajectories are identified. (iii) The 6-DOF motion of each
cluster is then estimated using 3-D pose optimization. (iv) Given a
pose estimate for each identified cluster, the most likely kinematic
structure and model parameters for the articulated object are
determined. Figure~\ref{fig:pipeline-text-training} illustrates the
steps involved in the training phase with inputs and outputs for each
component.

\begin{figure}[!h]
  \centering
    \includegraphics[width=0.95\columnwidth]{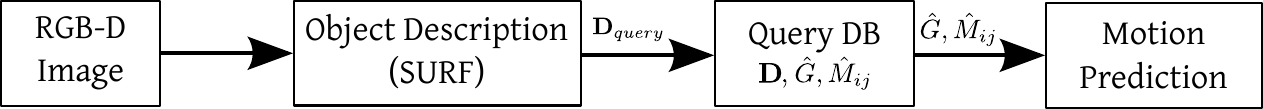}
  \caption{The prediction phase.}
  \label{fig:pipeline-text-prediction}
\end{figure}

Once the kinematic model of an articulated object is learned, our
system can predict the motion trajectory of the object during future
encounters. In the prediction phase: (i) Given RGB-D data, the
description of the objects in the scene, $\textbf{\text{D}}_{query}$,
is extracted using SURF~\cite{baya2008speeded} descriptors. (ii) Given
a set of descriptors $\textbf{\text{D}}_{query}$, the best-matching
object and its kinematic model, $\hat{G}, \hat{M}_{ij}, (ij) \in
\hat{G}$ are retrieved; and (iii) From these correspondences and the
kinematic model parameters of the matching object, the object's
articulated motion is
predicted. Figure~\ref{fig:pipeline-text-prediction} illustrates the
steps involved in the prediction phase.

\subsection{Spatio-Temporal Feature Tracking}
\label{sec:feature-tracking} 
The first step in articulation learning from visual demonstration
involves visually observing and tracking features on the object while
it is being manipulated. We focus on unstructured environments without
fiducial markers.  Our algorithm combines interest-point detectors and
feature descriptors with traditional optical flow methods to construct
long-range feature trajectories. We employ Good Features To Track
(GFTT)~\cite{shi1994good} to initialize up to 1500 salient features
with a quality level of 0.04 or greater, across multiple image
scales. Once the features are detected, we populate a mask image that
captures regions where interest points are detected at each
pyramid scale. We use techniques from previous work on dense optical
flow~\cite{farneback2003two} to predict each feature at the next
timestep. Our implementation also employs median filtering as
suggested by~\citet{wang2011action} to reduce false positives.




We bootstrap the detection and tracking steps with a feature
description step that extracts and learns the description of the
feature trajectory. At each image scale, we compute the SURF
descriptor~\cite{baya2008speeded} over features that were predicted
from the previous step, denoted as $\hat{f^t}$, and compare them with
the description of the detected features at time $t$, denoted as
$f^t$. Subsequently, detected features $f^t$ that are sufficiently
close to predicted features $\hat{f^t}$ and that successfully meet a
desired match score are added to the feature trajectory, while the
rest are pruned. To combat drift, we use the detection mask as a guide
to reinforce feature predictions with feature
detections. Additionally, we incorporate flow failure detection
techniques~\cite{kalal2010forward} to reduce drift in feature
trajectories.



Like other feature-based methods~\cite{katz2013interactive} our method
requires visual texture.  In typical video sequences, some features
are continuously tracked, while other features are lost due to
occlusion or lack of image saliency. To provide rich trajectory
information, we continuously add features to the scene as needed. We
maintain a constant number of feature trajectories tracked, by adding
newly detected features in regions that are not yet occupied. From
RGB-D depth information, image-space feature trajectories can be
easily extended to 3-D. As a result, each feature key-point is
represented by its normalized image coordinates $(u,v)$, position
$\vec{p} \in \mathbb{R}^3$ and surface normal $\vec{n}$, represented
as $(\vec{p},\vec{n}) \in \mathbb{R}^3 \times SO(2)$. We denote
$\textbf{\text{F}} = \{F_1, \dots, F_n\}$ as the resulting set of
feature trajectories constructed, where $F_i = \{(\vec{p_1},
\vec{n_1}), \dots, (\vec{p_t}, \vec{n_t})\}$. To combat noise inherent
in our consumer-grade RGB-D sensor, we post-process the point cloud
with a fast bilateral filter~\cite{paris2006fast} with parameters
$\sigma_s=20$~px, $\sigma_r=4$~cm.

\subsection{Motion Segmentation} 
%
To identify the kinematic relationships among parts in an articulated
object, we first distinguish the trajectory taken by each part.  In
particular, we analyze the motions of the object parts with respect to
each other over time, and infer whether or not pairs of object parts 
are rigidly attached. To reason over candidate segmentations, we formulate
a clustering problem to identify the different motion subspaces in
which the object parts lie. After clustering, similar labels imply
rigid attachment, while dissimilar labels indicate non-rigid relative 
motion between parts.

\begin{figure}[t!]
  \centering
    \includegraphics[width=0.95\columnwidth, clip, trim=0mm 95mm 0mm 0mm]{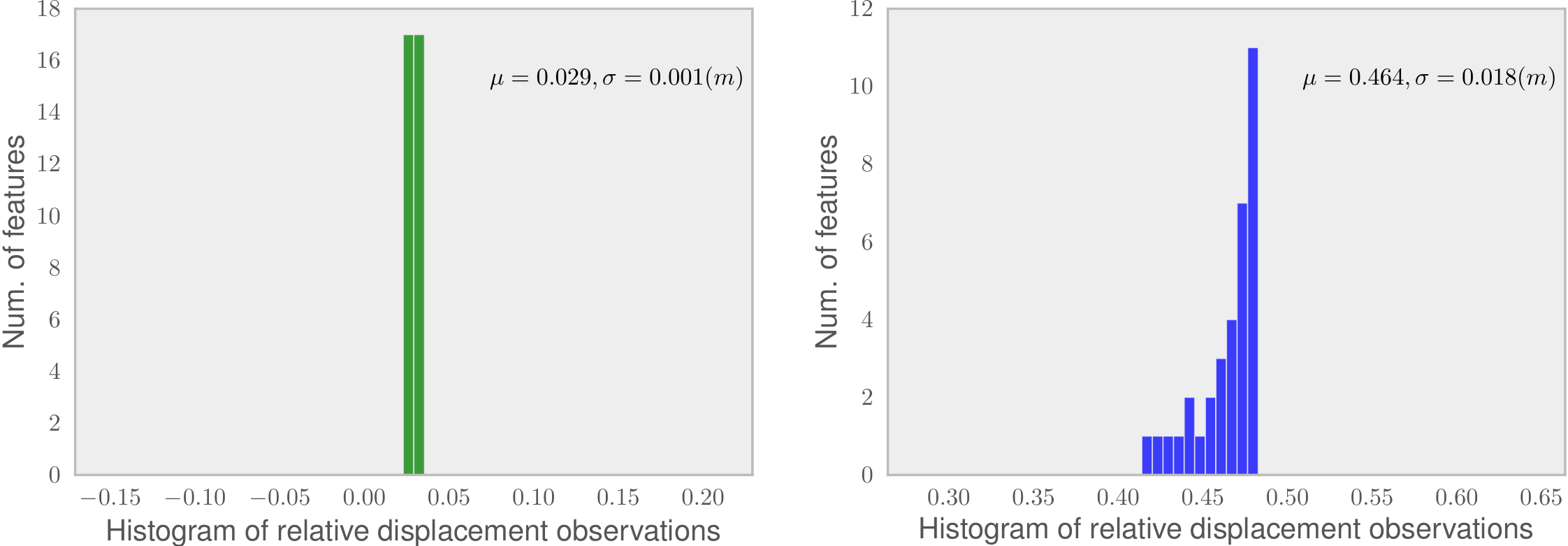} 
    \caption{Histogram of observed distances between a pair of
      trajectories accumulated over one demonstration. (Left) The
      distribution of observed distances is centered at $\mu=0.029~$m
      with $\sigma=0.001~$m, indicating rigid-body motion. (Right)
      Larger variation in observed distances, with $\sigma=0.018~$m,
      indicates non-rigid motion.}
  \label{fig:pose-pair-histogram}
\end{figure}

If two features in $\mathbb{R}^3 \times SO(2)$ belong to the same
rigid part, the relative displacement and angle between the features
will be consistent over the common span of their trajectories. The
distribution over the relative change in displacement vectors and
angle subtended is modeled as a zero-mean Gaussian,
\mbox{$\mathcal{N}(\mu,\Sigma) = (0, \Sigma)$}, where $\Sigma$ is the
expected noise covariance for rigidly-connected feature
pairs. The similarity of two feature trajectories can then be defined as:
%
%
\begin{equation}
L(i,j) = \frac{1}{T}\sum_{t \in t_i \cap t_j} \exp
\Bigg\{-\gamma ~ \Big(d(x_i^{t},x_j^{t})- \mu_{d_{ij}}\Big)^2\Bigg\}
\label{eqn:pairwise-affinity}
\end{equation}
where $t_i$ and $t_j$ are the observed time instances of the feature
trajectories $i$, and $j$ respectively, $T = \left\vert{t_i \cap
    t_j}\right\vert$, and $\gamma$ is a parameter characterizing the
relative motion of the two trajectories. For a pair of 3-D key-point
features $\vec{p_i}$, and $\vec{p_j}$, we estimate the mean relative
displacement between a pair of points moving rigidly together as:
\begin{equation}\label{eqn:dist-l2}
    \mu_{d_{ij}}=\frac{1}{T}\sum_{t \in t_i \cap t_j}d(\vec{p_i}^t,\vec{p_j}^t)
\end{equation}
where $d(\vec{p_i},\vec{p_j})=\|\vec{p_i} - \vec{p_j}\|$. For 3-D key-points, we
use $\gamma = \frac{1}{2~\text{cm}}$ in
Eqn.~\ref{eqn:pairwise-affinity}. Figure~\ref{fig:pose-pair-histogram}
illustrates an example of rigid and non-rigid motions of feature trajectory
pairs, and their corresponding distribution of relative displacements.

For a pair of surface normals
$\vec{n_i}$ and $\vec{n_j}$, we define the mean distance as
\begin{equation}
\mu_{d_{ij}}=\frac{1}{T}\sum_{t \in t_i \cap t_j}d(\vec{n_i}^t,\vec{n_j}^t),
\label{eqn:dist-cosine}
\end{equation}
where $\quad d(\vec{n_i},\vec{n_j})=1-\vec{n_i} \cdot
\vec{n_j}$. In this case, we use \mbox{$\gamma =
\frac{1}{\cos(15\,^{\circ})}$} in
Eqn.~\ref{eqn:pairwise-affinity}. 

%


Since the bandwidth parameter $\gamma$ for a pair of feature trajectories can be
intuitively predicted from the expected variance in relative motions of
trajectories, we employ DBSCAN~\citep{ester1996density}, a density-based
clustering algorithm, to find rigidly associated feature trajectories. The
resulting cluster assignments are denoted as $\textbf{\text{C}} = \{C_1, \dots,
C_k\}$, where cluster $C_i$ consists of a set of rigidly-moving feature
trajectories. 




\subsection{Multi-Rigid-Body Pose Optimization}
Given the cluster label assignment for each feature trajectory, we
subsequently determine the 6-DOF motion of each cluster. We define $Z_i^t$ as
the set of features belonging to cluster $C_i$ at time $t$. Additionally, we
define $\textbf{X} = {X_1, \dots, X_k}$ as the set of $SE(3)$ poses estimated
for each of $k$ clusters considered, and $x_i^t \in X_i$ as the $SE(3)$ pose
estimated for the $i^{th}$ cluster at time $t$.



For each cluster $C_i$, we consider the synchronized sensor
observations of position and surface normals for each of its
trajectories, and use the arbitrary pose $x_i^0$ as the reference
frame for the remaining pose estimates of the $i^{th}$
cluster. Subsequently, we compute the relative transformation
$\Delta_i^{t-1,t}$ between successive time steps $t-1$ and $t$ for the
$i^{th}$ cluster using the known correspondences between $Z_i^{t-1}$
and $Z_i^t$. Since this step can lead to drift, we add an additional
sparse set of relative pose constraints every 10 frames, denoted as
$\Delta_i^{t-10,t}$. Our implementation employs a correspondence
rejection step that eliminates outliers falling outside the inlier
distance threshold of $1$~cm, as in RANSAC~\cite{fischler1981random},
making the pose estimation routine more robust to sensor noise.


We augment the estimation step with an optimization phase to provide
smooth and continuous pose estimates for each cluster by incorporating
a motion model. We use the 3-D pose optimizer
iSAM~\cite{kaess2012isam2} to incorporate the relative pose
constraints within a factor graph, with node factors derived directly
from the pose estimates.  A constant-velocity edge factor term is also
added to provide continuity in the articulated motion.



\subsection{Articulation Learning} 
Once the 6-DOF pose estimates of the individual object parts are computed, the
kinematic model of the full articulated object is determined using
tools developed in ~\citet{sturm2011probabilistic}. Given multiple 6-DOF pose
observations of object parts, the problem is to estimate the most likely
kinematic configuration for the articulated object. Formally, given the
observed poses $\mathcal{D}_z$, we estimate the kinematic
graph configuration $\hat{G}$ that maximizes the posterior probability
\begin{equation} \hat{G} = \underset{G}{\arg\max}~p(G\mid\mathcal{D}_z)
\label{eq:posterior-prob-max}
\end{equation}

We employ notation similar to that of~\citet{sturm2011probabilistic} to
denote the relative transformation between two object parts $i$ and $j$ as
$\Delta_{ij} = x_i~\ominus~x_j$, using standard motion composition operator
notation~\cite{smith1990estimating}. The kinematic model between part $i$ and
$j$ is then defined as $M_{ij}$, with its associated parameter vector
$\theta_{ij} \in \mathbb{R}^{p_{ij}}$, where $p_{ij}$ are the number of
parameters associated with the description of the link. We construct a graph $G
= (V_G,E_G)$ consisting of a set of vertices $V_G = {1, \dots, k}$ that denote
the object parts involved in the articulated object, and a set of undirected
edges $E_G \subset V_G \times V_G$ describing the kinematic linkage between two
object parts.




As in \citet{sturm2011probabilistic}, we simplify the problem to
recognize only kinematic trees of high posterior probability, 
in order to reformulate the problem as equation~\ref{eq:posterior-prob-tree} below: 
\begin{align} \hat{G} &= \underset{G}{\arg\max}~p(G\mid\mathcal{D}_z)\\ 
&=
\underset{G}{\arg\max}~p(\left\{(\mathcal{M}_{ij},\theta_{ij}) \mid (ij) \in
E_G\right\}\mid\mathcal{D}_z)\\ 
&= \underset{G}{\arg\max}~\prod_{(ij) \in
E_G}~p(\mathcal{M}_{ij},\theta_{ij}\mid\mathcal{D}_z)\\
&= \underset{E_G}{\arg\max}~\sum_{(ij) \in E_G}~\log~p(\hat{\mathcal{M}_{ij}},\hat{\theta_{ij}}\mid\mathcal{D}_z)
\label{eq:posterior-prob-tree}
\end{align} where $\mathcal{D}_z = (\Delta_{ij}^1, \dots, \Delta_{ij}^t)~\forall~(ij) \in E_G$ is the
sequence of observed relative transformations between parts $i$ and $j$.

Since we are particularly interested in household objects, we focus on
kinematic models involving rigid, prismatic, and revolute linkages. We
then estimate the parameters $\theta \in \mathbb{R}^p$ that maximize
the data likelihood of the object pose observations given the
kinematic model:
\begin{align} \hat{\theta} &=
\underset{\theta}{\arg\max}~p(\mathcal{D}_z\mid\mathcal{M},\theta)
\label{eq:data-likelihood}
\end{align} 




Once we fit each candidate kinematic model
to the given observation sequence, we select the kinematic model that best
explains the data. Specifically, we compute the posterior probability of
each kinematic model, given the data, as:
\begin{align} ~p(\mathcal{M}\mid\mathcal{D}_z) =
\bigintsss\frac{p(\mathcal{D}_z\mid\mathcal{M},\theta)~p(\theta\mid\mathcal{M})~p(\mathcal{M})}{p(\mathcal{D}_z)}
d\theta
\label{eq:posterior-prob-structure-selection}
\end{align}

Due to the evaluation complexity of this posterior term, the BIC score is
computed instead as the approximation: 
\begin{align} BIC(\mathcal{M}) =
-2\log~p(\mathcal{D}_z\mid\mathcal{M},\hat{\theta}) + p\log n,
\label{eq:bic-structure-selection}
\end{align}where $p$ is the number of parameters
involved in the kinematic model, $n$ is the number of observations in the
data set, and $\hat{\theta}$ is the maximum likelihood parameter
vector. This implies that the model that best explains the observations would
correspond to that with the least BIC score.



\begin{figure}[H]
  \centering
  \subfigure[Rotational DOF of a laptop]{\includegraphics[height=1.3in]{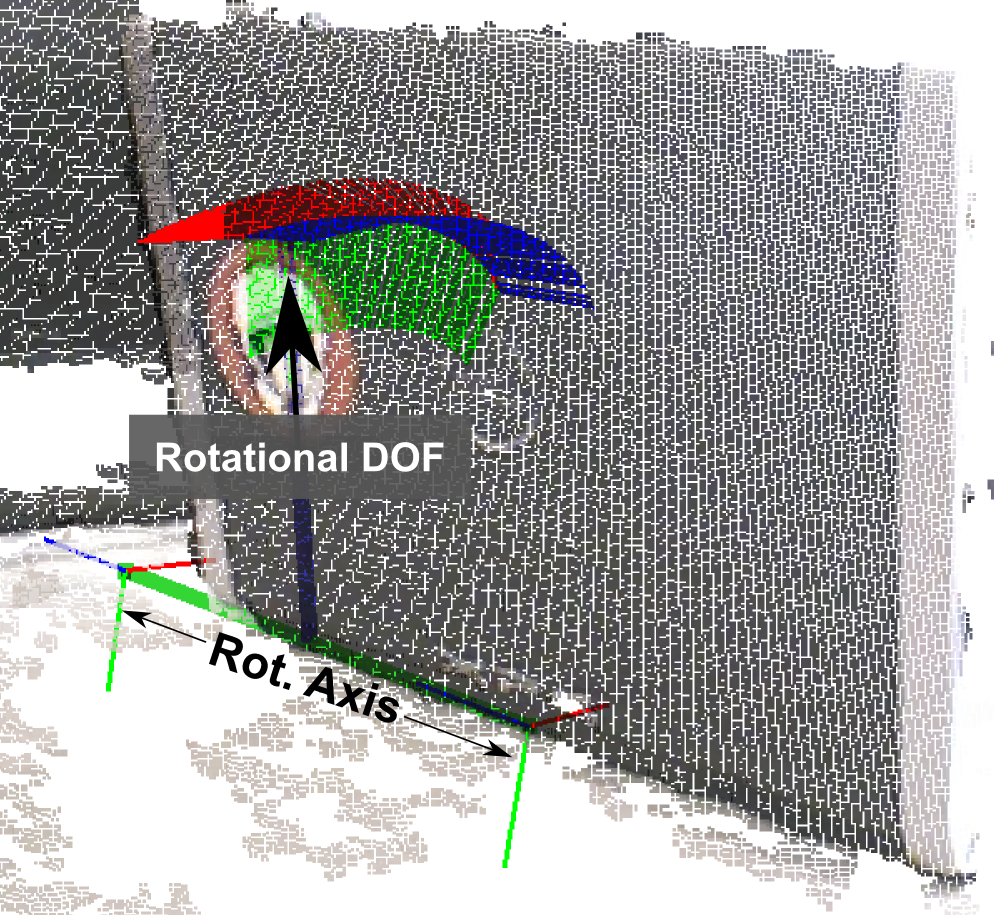}}~
  \subfigure[Prismatic DOF of a drawer]{\includegraphics[height=1.2in]{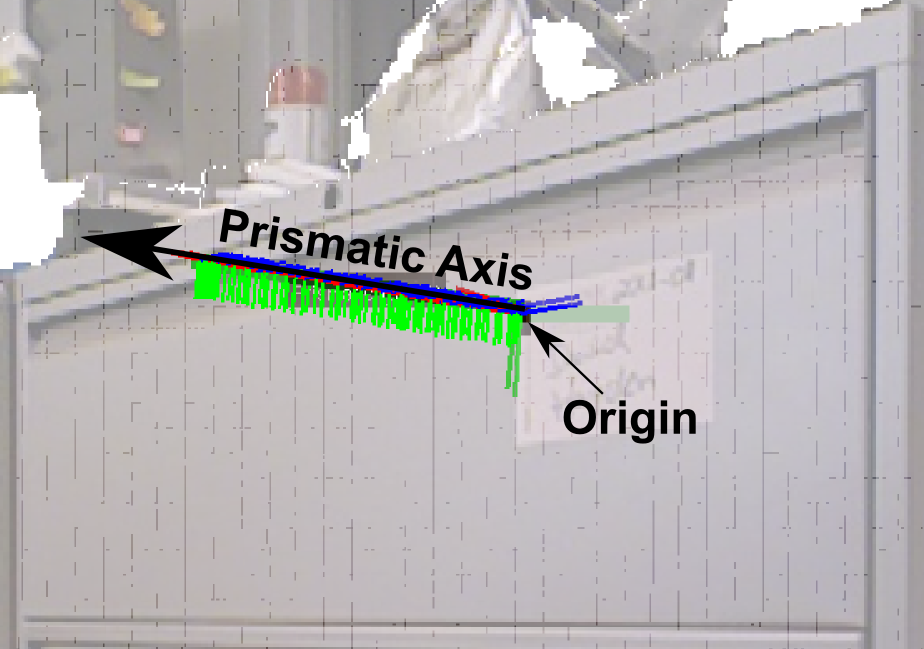}}
  \caption{Examples of correctly estimated kinematic structure from 6-DOF pose estimates
    of feature trajectories.}
  \label{fig:model-estimation}
\end{figure}

The kinematic structure selection problem is subsequently reduced to
computing the minimum spanning tree of the graph with edges defined by
$cost_{ij} =
-\log~p(\mathcal{M}_{ij},\theta_{ij}\mid\mathcal{D}_{z_{ij}})$. The
resulting minimum spanning kinematic tree weighted by BIC scores is
the most likely kinematic model for the articulated object given the
pose observations.  For a more detailed description, we refer the
reader
to~\citet{sturm2011probabilistic}. Figure~\ref{fig:model-estimation}
shows a few examples of kinematic structures extracted given pose
estimates as described in the previous section.  Our limitation of
linkage types to rigid, prismatic, and rotational does exclude various
household objects such as lamps, garage doors, toys etc.~with more
complex kinematics.

%


\subsection{Learning to Predict Articulated Motion}
Our daily environment is filled with articulated objects with
which we repeatedly interact. A robot in our environment can
identify instances of articulated objects that it has observed
in the past, then use a learned model to predict the motion 
of an object when it is used.

%
\begin{figure}[H]
  \centering
  \subfigure[Extracted MSER]{\includegraphics[width=0.5\columnwidth]{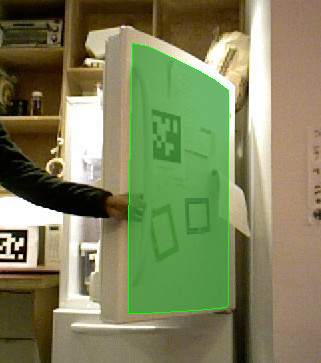}}~
  \subfigure[Estimated Motion Manifold]{\includegraphics[width=0.5\columnwidth]{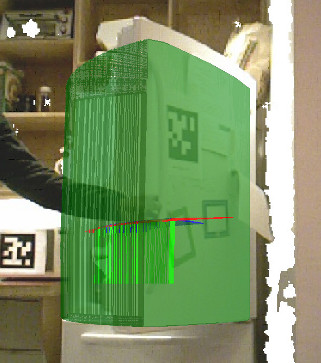}}

  \caption{The motion manifold of an articulated object, extracted via MSERs.}
  \label{fig:mser-surface}
\end{figure}

%
\begin{figure*}[!t]
  \centering
    \includegraphics[width=2\columnwidth]{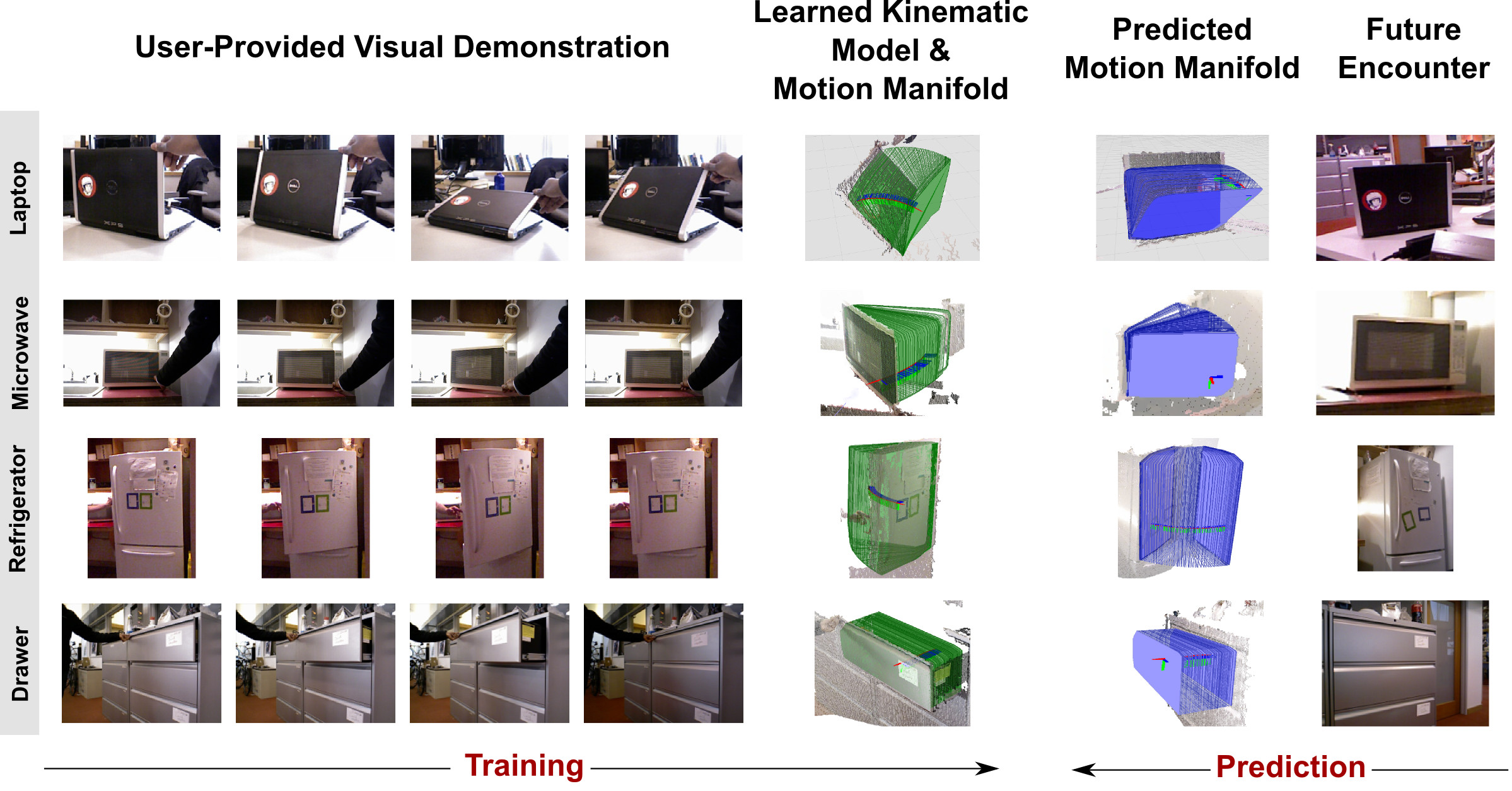}\\
  \caption{Articulation learning and motion prediction for various objects.}
  \label{fig:results-fig}
\end{figure*}

Once the kinematic model of an articulated object is learned, the kinematic
structure $\hat{G}$ and its model parameters $\hat{\mathcal{M}_{ij}}, (ij) \in
\hat{G}$ are stored in a database, along with its appearance model. The feature
descriptors extracted (described in Section~\ref{sec:feature-tracking}) for each
cluster $C_i$ of the articulated object are also retained for object recognition
in future encounters. Demonstrations involving the same instance of the
articulated object are represented in a single arbitrarily selected reference
frame, and kept consistent across encounters by registering
newer demonstrations into the initial object frame. Each of these
attributes is stored in the bag-of-words driven database~\cite{GalvezTRO12} for convenient querying in the
future. Thus, on encountering the same object instance in the future, the robot
can match the descriptors extracted from the current scene with those extracted
from object instances it learned in the past. It then recovers the original
demonstration reference frame along with the relevant kinematic structure of the
articulated object for prediction purposes. We identify the
surface of the manipulated object by extracting Maximally Stable Extremal
Regions (MSER)~\cite{matas2004robust} (Figure~\ref{fig:mser-surface})
for each object part undergoing motion. We use this surface to
visualize the motion manifold of the articulated object.

\section{Experiments and Analysis}



Our experimental setup consists of a single sensor providing RGB-D
depth imagery. Each visual demonstration involved a human manipulating
an articulated object and its parts at a normal pace, while avoiding
obscuration of the object from the robot's perspective.
Demonstrations were performed for multiple robot viewpoints, to
capture variability in depth imagery. We performed 43 demonstration
sessions by manipulating a variety of household objects:
refrigerators, doors, drawers, laptops, chair etc. Each demonstration was
recorded for about 30-60 seconds. April tags~\cite{olson2011tags} were
used to recover ground truth estimates of each articulated object's
motion, which we adopted as a baseline for evaluation. In order to
avoid any influence on our method of observations arising from
fiducial markers, the RGB-D input was pre-processed to mask out
regions containing the tags.

We then compared the pose estimation, model selection and estimation
performance of our method to that of an alternative state-of-the-art
method (re-implemented by us based on~\cite{katz2013interactive}), and
to traditional methods using fiducial markers.  We incorporated
several improvements~\cite{kalal2010forward},~\cite{paris2006fast} to
Katz's algorithm, as previously described in
Section~\ref{sec:feature-tracking}, 
to enable fair comparison with our proposed method.

\subsection{Qualitative and Overall Performance}
Figure~\ref{fig:results-fig} shows the method in operation for
household objects including a laptop, a microwave, a refrigerator and
a drawer. Tables~\ref{tab:pose-estimation-comparison} and~\ref{tab:model-estimation-comparison} compare the performance of
our method in estimating the kinematic model parameters for several
articulated objects observed from a variety of viewpoints.  Our method
recovered a correct model for more objects, and for almost every
object tested recovered model parameters more accurately, than Katz's
method.

\subsection{Pose Estimation Accuracy}
\begin{figure}[H]
  \centering
    \includegraphics[width=\columnwidth]{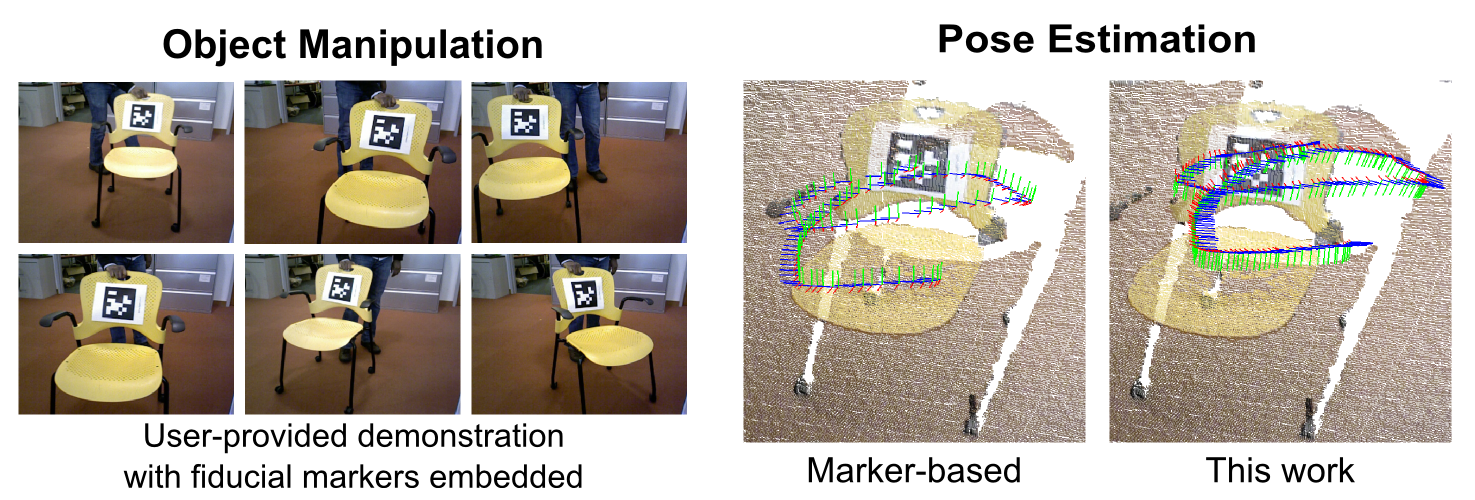}\\
  \caption{Pose estimation accuracy of our method, compared 
    to that achieved using fiducial markers.}
  \label{fig:pose-estimation-eval}
\end{figure}

For each visual demonstration, we compared the segmentation and
$SE(3)$ pose of each object part estimated by our method with those
produced by Katz.  We also obtained pose estimates for each object
part by tracking attached fiducial markers.  Synchronization across
pose observations was ensured by evaluating only poses in the set
intersection of the timestamps of each pose sequence. For each
overlapping time step, we compared the relative pose of the estimated
object segment obtained from both algorithms with that obtained via
fiducial markers (Figure~\ref{fig:pose-estimation-eval}). For
consistency in evaluation, the $SE(3)$ poses of individual object
parts were initialized identically for both algorithms.

\begin{figure*}[t]
  \centering
  \subfigure[Accurate estimation by current state-of-the-art and our framework]{\includegraphics[width=\columnwidth]{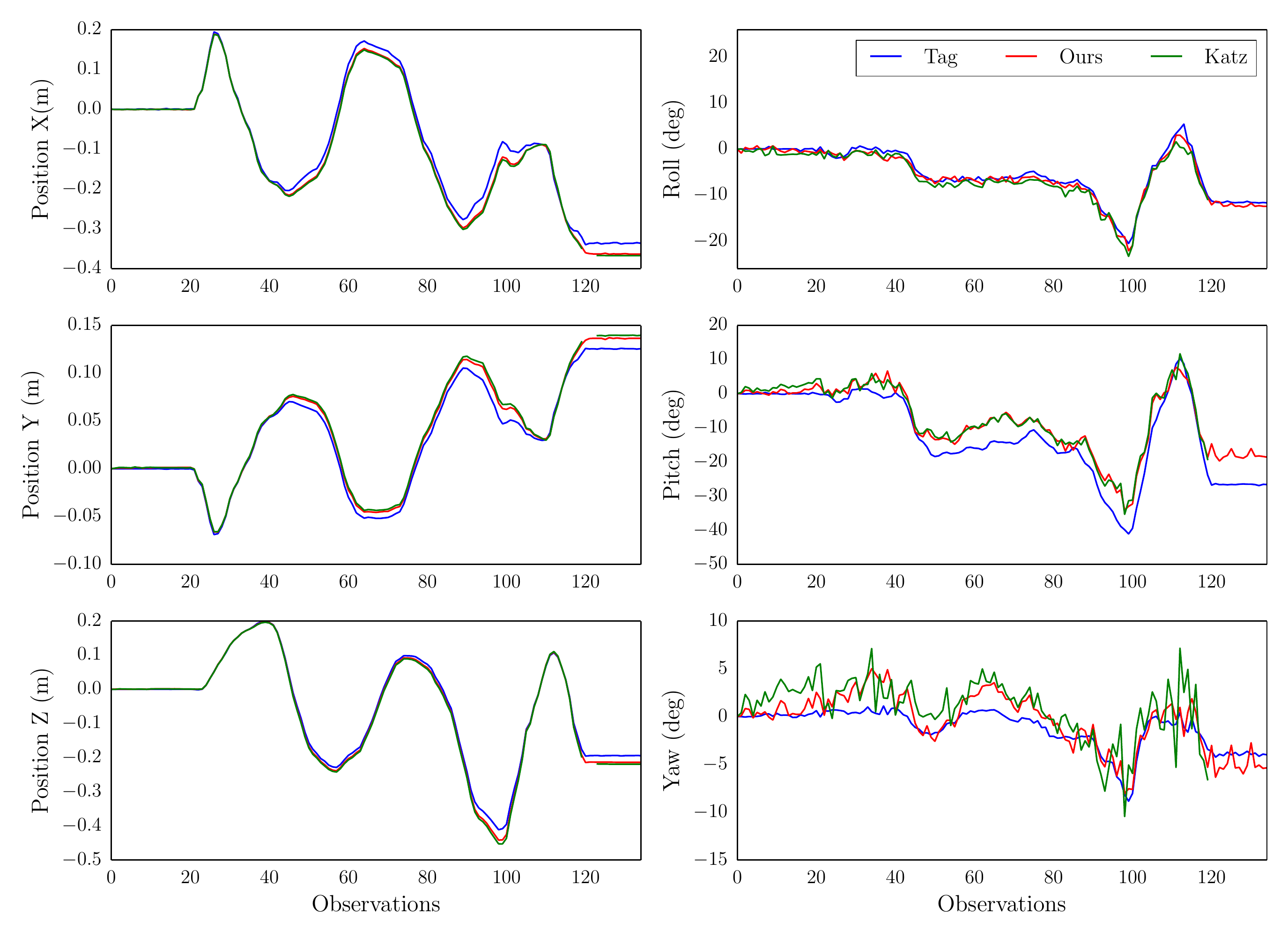}\label{fig:chair-accurate}}
  \subfigure[Failed estimation by current state-of-the-art
  ]{\includegraphics[width=\columnwidth]{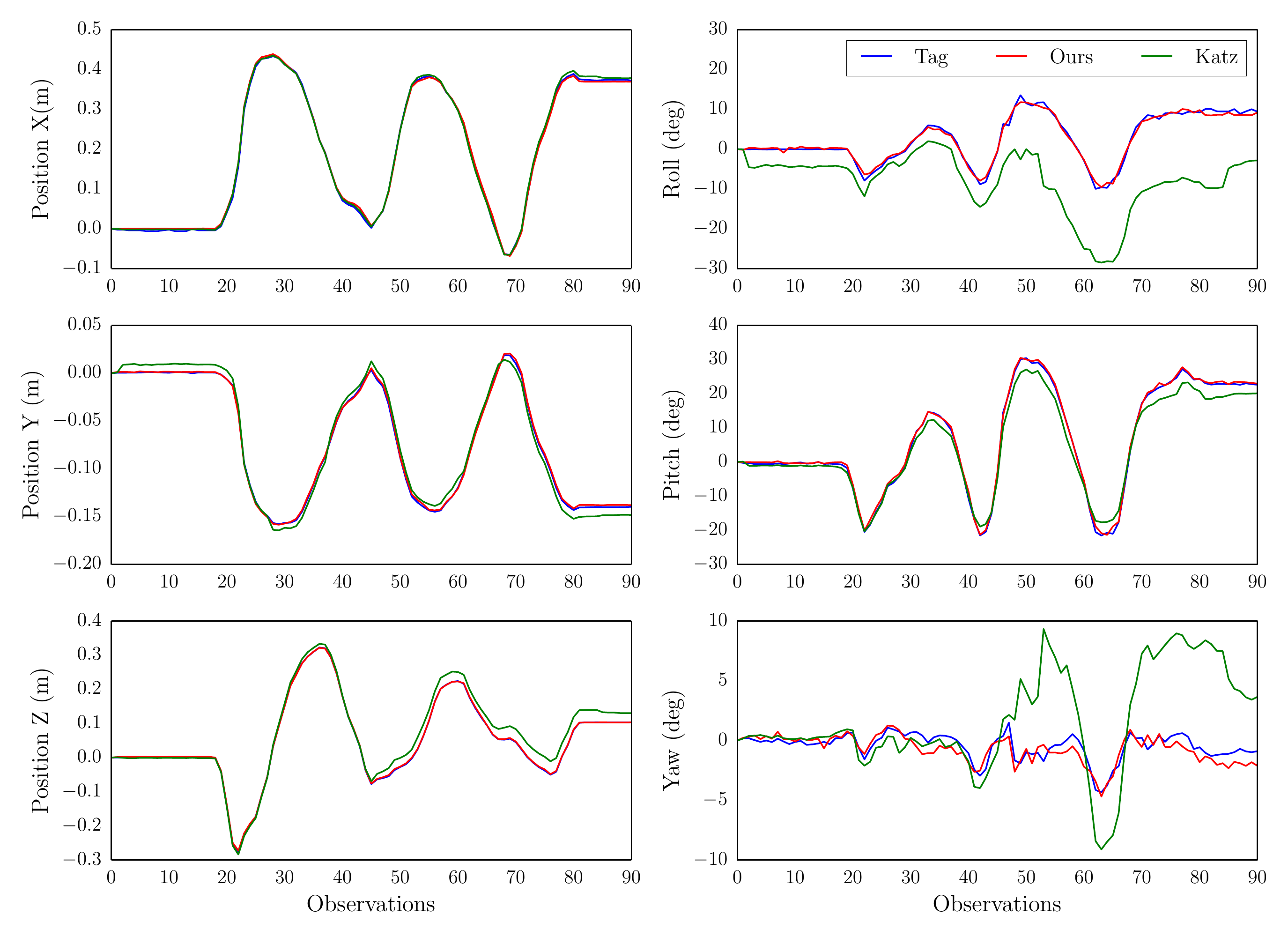}\label{fig:chair-fail}}

  \caption{Comparison of $SE(3)$ pose for a chair estimated via
    fiducial markers (Tag), current state-of-the-art (Katz) and our framework
    (Ours). (a) The figures show the strong performance of our framework, as
    compared to marker-based solutions and current
    state-of-the-art algorithms, to robustly track and
    estimate the $SE(3)$ pose of a chair being manipulated on multiple
    occasions. (b) Current state-of-the-art, however, fails to robustly
    estimate the $SE(3)$ pose on certain trials. }
  \label{fig:chair-motion-accuracy-comparison}
\end{figure*}

Figure~\ref{fig:chair-motion-accuracy-comparison} compares the
absolute $SE(3)$ poses estimated by the three methods described above,
given observations of a chair being moved on the ground plane.
Figure~\ref{fig:chair-accurate} illustrates a scenario in which both
algorithms, ours and Katz's, perform reliably.  Katz's method is
within $2.0~$cm and $2.6\,^{\circ}$, on average, of the ground truth pose produced
with fiducial markers.  Our method achieves comparable average accuracy of
$1.7~$cm and $2.1\,^{\circ}$.  Using data from another demonstration,
Katz's method failed to track the object motion robustly, resulting in
drift and incorrect motion estimates
(Figure~\ref{fig:chair-fail}). Such failures can be attributed
to: \textit{(i)} the KLT tracker that is known to cause drift 
during feature tracking; \textit{(ii)} SVD least squares minimization
in the relative pose estimation stage, without appropriate outlier
rejection.

For a variety of articulated objects
(Table~\ref{tab:pose-estimation-comparison}), our method achieves
average accuracies of $2.4~$cm and $4.7\,^{\circ}$ with respect to
ground truth estimated from noisy Kinect RGB-D data. In comparison,
Katz's method~\cite{katz2013interactive} achieved average accuracies
of $3.7~$cm and $10.1\,^{\circ}$~for the same objects.  Our method
achieved an average error of less than $10~$cm and $25\,^{\circ}$
in 37 of 43 demonstrations, vs.~23 of 43 for Katz.

%

\subsection{Model Estimation Accuracy} Once the $SE(3)$ poses of the object
parts are estimated, we compare the kinematic structure and model parameters of
the articulated object estimated by our method with those produced by Katz. As
in our other experiments, we use the kinematic structure and model parameters
identified from fiducial marker-based solutions as a
baseline. Table~\ref{tab:model-estimation-comparison} summarizes the model
estimation and parameter estimation performance achieved with our method and
Katz's. The model fit error is defined as the average spatial and orientation
error between the $SE(3)$ observations and the estimated articulation manifold
(i.e.~prismatic or rotational manifold). For the dataset of articulated objects
evaluated (Table~\ref{tab:model-estimation-comparison}), our method achieved an
average model fit error of $1.7~$cm spatially, and $5.0\,^{\circ}$ in
orientation, an improvement over Katz's method (average model fit errors of
$2.0~$cm and $5.8\,^{\circ}$~respectively). Of 43 demonstrations evaluated, our
method determined the correct kinematic structure and accurate parameters in 30
cases, whereas Katz did so in only 15 cases. 

We also compared the model parameters estimated by our method and Katz's method
with ground truth from markers, by transforming poses estimated by both methods
into the fiducial marker's reference frame based on the initial configuration of
the articulated object. This allows us to directly compare model parameters
estimated through our proposed framework, the current state-of-the-art and
marker-based solutions. For multi-DOF objects, the model parameter error
averaged across each corresponding object part is reported. In each
demonstration, the model parameters estimated via our method are closer to the
marker-based solution than those obtained by Katz.

\begin{table*}[!thp]
  \centering
  \scriptsize
  \begin{minipage}[t]{0.85\columnwidth}
    \centering
    {\setlength{\tabcolsep}{0.4em}
        \begin{tabular}{|l|c|c|r|c|c|c|c|}
          \hline
          \multirow{3}{*}{\textbf{Dataset}} &
          \multirow{3}{*}{\rotatebox{90}{\textbf{DOF}}} & \multicolumn{ 3}{c|}{\textbf{Katz
              et al.}} & \multicolumn{ 3}{c|}{\textbf{Ours}} \\ \cline{ 3- 8}
          \multicolumn{ 1}{|c|}{} & \multicolumn{ 1}{c|}{} &\multicolumn{ 2}{c|}{\textbf{Average Error}}
          & 
          \multirow{2}{*}{\parbox{0.75cm}{\centering\textbf{Success Rate}}} & 
          \multicolumn{ 2}{c|}{\textbf{Average Error}} & 
          \multirow{2}{*}{\parbox{0.75cm}{\centering\textbf{Success Rate}}} \\
          \cline{ 3- 4}\cline{ 6- 7}
          
          \multicolumn{ 1}{|c|}{} & \multicolumn{ 1}{c|}{} &\textbf{Pos.} & \textbf{Orient.} &
          \multicolumn{ 1}{c|}{} &\textbf{Pos.} & \textbf{Orient.} & \multicolumn{ 1}{c|}{} \\ \hline
          \multicolumn{1}{|l|}{Door} & \multicolumn{ 1}{c|}{1} &6.0~cm & 6.8$\,^{\circ}$ & 6/7 & \textbf{5.0}~cm & \textbf{5.5}$\,^{\circ}$ &  \textbf{7/7} \\ 
          \multicolumn{1}{|l|}{Drawer} & \multicolumn{ 1}{c|}{1} & 6.1~cm & 18.0$\,^{\circ}$ & 3/7 & \textbf{3.7}~cm & \textbf{3.0}$\,^{\circ}$ & \textbf{6/7} \\ 
          \multicolumn{1}{|l|}{Fridge} & \multicolumn{ 1}{c|}{1} & 2.2~cm & 8.1$\,^{\circ}$ & 4/8 & \textbf{1.0}~cm & \textbf{2.9}$\,^{\circ}$ &  \textbf{6/8}\\ 
          \multicolumn{1}{|l|}{Laptop} & \multicolumn{ 1}{c|}{1} & 0.4~cm & 2.3$\,^{\circ}$ & 2/5 & \textbf{0.3}~cm & \textbf{6.4}$\,^{\circ}$ &  \textbf{4/5}\\ 
          \multicolumn{1}{|l|}{Microwave} & \multicolumn{ 1}{c|}{1} & 4.3~cm & 14.2$\,^{\circ}$ & 2/4 & \textbf{1.9}~cm & \textbf{6.9}$\,^{\circ}$ & \textbf{4/4} \\ 
          \multicolumn{1}{|l|}{Printer} & \multicolumn{ 1}{c|}{1} & 0.7~cm & 2.5$\,^{\circ}$ & \textbf{1/2} & \textbf{0.5}~cm & \textbf{2.3}$\,^{\circ}$ &  \textbf{2/2}\\ 
          \multicolumn{1}{|l|}{Screen} & \multicolumn{ 1}{c|}{1} & \textbf{2.6}~cm & 24.9$\,^{\circ}$ & 1/2 & 3.4~cm &
          \textbf{3.5}$\,^{\circ}$ &  \textbf{1/2}\\ 

          \multicolumn{1}{|l|}{Chair} & \multicolumn{ 1}{c|}{2} &3.6~cm & 13.2$\,^{\circ}$ & 2/3 & \textbf{2.3}~cm & \textbf{4.5}$\,^{\circ}$ & \textbf{3/3} \\ 
          \multicolumn{1}{|l|}{Monitor} & \multicolumn{ 1}{c|}{2} &\textbf{0.8}~cm & 7.2$\,^{\circ}$ & 1/2 & 1.8~cm &
          \textbf{2.3}$\,^{\circ}$ &  \textbf{2/2}\\  
          \multicolumn{1}{|l|}{Bicycle} & \multicolumn{ 1}{c|}{3} &1.7~cm & 10.4$\,^{\circ}$ & 1/3 & \textbf{1.1}~cm &
          \textbf{9.8}$\,^{\circ}$ &  \textbf{2/3}\\ \hhline{|=|=|=|=|=|=|=|=|}

          \multicolumn{1}{|l}{\textbf{Overall}} & \multicolumn{ 1}{c|}{} &3.7~cm & 10.1$\,^{\circ}$ & 23/43 & \textbf{2.4}~cm &
          \textbf{4.7}$\,^{\circ}$ &  \textbf{37/43}\\ \hline
        \end{tabular}
      }
\captionof{table}{Comparison of $SE(3)$ pose estimates between our framework
  and current state-of-the-art (Katz) with marker-based pose estimates
  considered as ground truth. 
}
\label{tab:pose-estimation-comparison}
\end{minipage}\hfill
\begin{minipage}[t]{1.15\columnwidth}
  \centering
  {\setlength{\tabcolsep}{0.4em}
    \centering
    \begin{tabular}{|l|c|c|c|c|c|c|c|c|c|}
      \hline

          \multirow{3}{*}{\textbf{Dataset}} & \multirow{3}{*}{\rotatebox{90}{\textbf{DOF}}} & \multicolumn{ 4}{c|}{\textbf{Katz
              et al.}} & \multicolumn{ 4}{c|}{\textbf{Ours}} \\ \cline{ 3- 10}
          \multicolumn{1}{|c|}{} & \multicolumn{ 1}{c|}{} & \multicolumn{ 2}{c|}{\textbf{Model Fit Error}} &
          \multirow{2}{*}{\parbox{1.2cm}{
              \centering{\renewcommand{\arraystretch}{1}\textbf{Param. Est. Error}}}} &
          \multirow{2}{*}{\parbox{0.75cm}{\centering\textbf{Success Rate}}} &
          \multicolumn{2}{c|}{\textbf{Model Fit Error}} &
          \multirow{2}{*}{\parbox{1.2cm}{\centering\textbf{Param. Est. Error}}} & 
          \multirow{2}{*}{\parbox{0.75cm}{\centering\textbf{Success Rate}}} \\ \cline{ 3- 4}\cline{ 7- 8}
          \multicolumn{1}{|c|}{} & \multicolumn{ 1}{c|}{} & \textbf{Pos.} & \textbf{Orient.} &
          \multicolumn{ 1}{c|}{} & \multicolumn{ 1}{c|}{} &
          \textbf{Pos.}& 
          \textbf{Orient.}& 
          \multicolumn{1}{c|}{} & \multicolumn{ 1}{c|}{} \\ \hline

          \multicolumn{1}{|l|}{Door} & \multicolumn{ 1}{c|}{1} & 1.9~cm & 6.7$\,^{\circ}$ & 1.9$\,^{\circ}$ & 4/7 & \textbf{0.4}~cm & \textbf{4.7}$\,^{\circ}$ & \textbf{1.8}$\,^{\circ}$ & \textbf{5/7} \\ 
          \multicolumn{1}{|l|}{Drawer} & \multicolumn{ 1}{c|}{1} & 2.0~cm & 7.3$\,^{\circ}$ & 2.5$\,^{\circ}$ & 2/7 & \textbf{1.7}~cm &
          \textbf{3.1}$\,^{\circ}$ & \textbf{2.0}$\,^{\circ}$ & \textbf{6/7} \\ 
          \multicolumn{1}{|l|}{Fridge} & \multicolumn{ 1}{c|}{1} & 0.5~cm & 6.5$\,^{\circ}$ & 5.6$\,^{\circ}$ & 4/8 & \textbf{0.4}~cm &
          \textbf{5.8}$\,^{\circ}$ & \textbf{3.5}$\,^{\circ}$ & \textbf{5/8}\\ 
          \multicolumn{1}{|l|}{Laptop} & \multicolumn{ 1}{c|}{1} & - & - & - & 0/5 & \textbf{0.2}~cm &
          \textbf{6.4}$\,^{\circ}$ & \textbf{6.1}$\,^{\circ}$ & \textbf{4/5}\\ 
          \multicolumn{1}{|l|}{Microwave} & \multicolumn{ 1}{c|}{1} & 7.0~cm & \textbf{1.2}$\,^{\circ}$ & \textbf{0.2}$\,^{\circ}$ & 2/4 & \textbf{6.5}~cm &
          4.1$\,^{\circ}$ & 0.3$\,^{\circ}$ & \textbf{3/4} \\ 
          \multicolumn{1}{|l|}{Printer} & \multicolumn{ 1}{c|}{1} & \textbf{0.9}~cm & 0.8$\,^{\circ}$ & 1.5$\,^{\circ}$ & 1/2 & 2.1~cm &
          \textbf{0.2}$\,^{\circ}$ & \textbf{1.4}$\,^{\circ}$ & \textbf{1/2}\\ 
          \multicolumn{1}{|l|}{Screen} & \multicolumn{ 1}{c|}{1} & - & - & - & 0/2 & \textbf{0.9}~cm &
          \textbf{0.7}$\,^{\circ}$ & \textbf{3.2}$\,^{\circ}$ & \textbf{1/2}\\  

          \multicolumn{1}{|l|}{Chair} & \multicolumn{ 1}{c|}{2} & \textbf{0.3}~cm & 11.2$\,^{\circ}$~ & 9.8$\,^{\circ}$ & 1/3 & 3.9~cm &
          \textbf{7.9}$\,^{\circ}$ & \textbf{4.8}$\,^{\circ}$ & \textbf{2/3}\\ 

          \multicolumn{1}{|l|}{Monitor} & \multicolumn{ 1}{c|}{2} & - & - & - & 0/2 & \textbf{2.9}~cm &
          \textbf{6.4}$\,^{\circ}$ & \textbf{5.7}$\,^{\circ}$ & \textbf{1/2}\\ 

          \multicolumn{1}{|l|}{Bicycle} & \multicolumn{ 1}{c|}{3} & 0.9~cm & \textbf{5.1}$\,^{\circ}$ & \textbf{4.2}$\,^{\circ}$ & 1/3 & 0.7~cm &
          8.5$\,^{\circ}$ & 7.3$\,^{\circ}$ & \textbf{2/3}\\ \hhline{|=|=|=|=|=|=|=|=|=|=|}

          \multicolumn{1}{|l}{\textbf{Overall}} & \multicolumn{ 1}{c|}{} & 2.0~cm & 5.8$\,^{\circ}$ & 3.4$\,^{\circ}$ & 15/43 & \textbf{1.7}~cm &
          \textbf{5.0}$\,^{\circ}$ & \textbf{3.3}$\,^{\circ}$ & \textbf{30/43}\\ \hline
        \end{tabular}
      }
      \captionof{table}{Comparison of kinematic model estimation and parameter
        estimation capability between our framework
  and current state-of-the-art (Katz) with marker-based model estimation 
  considered as ground truth. 
}
      \label{tab:model-estimation-comparison}
    \end{minipage}
\end{table*}





\section{Conclusion} 
We introduced a framework that enables robots to learn kinematic
models for everyday objects from RGB-D data acquired during
user-provided demonstrations. We combined sparse feature tracking,
motion segmentation, object pose estimation and articulation learning
to learn the underlying kinematic structure of the observed object.
We demonstrated the qualitative and quantitative performance of our
method; it recovers the correct structure more often, and more
accurately, than its predecessor in the literature, and achieves
accuracy similar to that of a marker-based solution.  Our framework
also enables the robot to predict the motion of articulated objects it
has previously learned.  Even given our method's limitation to
recovering open kinematic chains involving only rigid, prismatic or
revolute linkages, its prediction capability may be useful in
future robotic encounters requiring manipulation.




\bibliographystyle{abbrvnat} 
\bibliography{references}

\end{document}